%% file: 2025_arr_GTD.tex
\pdfoutput=1

\documentclass[11pt]{article}
\usepackage{acl}

\newif\ifcomment\commenttrue
\input{style/preamble}

\newcommand{\gtd}{\abr{gtd}}
\newcommand{\globalterrorism}{Global Terrorism Database}

\newcommand{\incidentset}[1]{%
  \ifnum#1=1 
    event set%
  \else
    \ifnum#1=2
      Event set
    \else
      Event Set
    \fi
  \fi
}

\newcommand{\cls}{\abr{llm\nobreakdash-cls}}
\newcommand{\emb}{\abr{embedding}}
\newcommand{\seg}{\abr{llm\nobreakdash-cls+seg}}
\newcommand{\cdcr}{Event Set Curation}
\newcommand{\cdeae}{Variable Coding}

\usepackage{times}
\usepackage{latexsym}
\usepackage[T1]{fontenc}
\usepackage[utf8]{inputenc}
\usepackage{microtype}
\usepackage{inconsolata}
\usepackage{graphicx}
\usepackage{booktabs}
\usepackage{amsmath}
\usepackage{siunitx}
\usepackage{longtable}

\usepackage{algorithm}
\usepackage{algpseudocode}

\title{
Large Language Models Are Effective Human Annotation Assistants, But Not Good Independent Annotators
}

\author{Feng Gu$^{1}$ \hspace{0.3cm} Zongxia Li$^{1}$ \hspace{0.3cm} Carlos Rafael Colon$^{2}$ \\  \textbf{Benjamin Evans}$^{2}$ \hspace{0.3cm} \textbf{Ishani Mondal}$^{1}$ \hspace{0.3cm} \textbf{Jordan Lee Boyd-Graber}$^{1}$ \\
  $^{1}$Department of Computer Science, University of Maryland \\ 
    $^{2}$National Consortium for the Study of Terrorism and Responses to Terrorism \\
  \texttt{\{fgu1, zli12321, raffy, benevans, imondal, ying\}@umd.edu}\\
}

\begin{document}
\maketitle

\input{pages/00-abstract}

\input{pages/10-intro}

\input{pages/20-background}

\input{pages/30-incident_set_curation}

\input{pages/40-annotation}
\input{pages/50-related_work}
\input{pages/60-conclusion}

\input{pages/70-limitations}


\bibliography{bib/jbg, bib/2025_arr_GTD}

\input{pages/80-appendix}

\end{document}

%% file: style/preamble.tex
\usepackage[a-1b]{pdfx}

\usepackage{framed}
\usepackage{mdwlist}
\usepackage{siunitx}
\usepackage{latexsym}
\usepackage{colortbl}
\usepackage{xcolor}
\usepackage{nicefrac}
\usepackage{booktabs}
\usepackage{fnpct}
\usepackage{amsfonts}
\usepackage[T1]{fontenc}
\usepackage{bold-extra}
\usepackage{amsmath}
\usepackage{amssymb}
\usepackage{bm}
\usepackage{graphicx}
\usepackage{mathtools}
\usepackage{microtype}
\usepackage{multirow}
\usepackage{multicol}
\usepackage{xpatch}
\usepackage{latexsym,comment}
\usepackage[normalem]{ulem}

\newcommand*{\missingreference}{{\Huge \colorbox{red}{?reference?}}}
\newcommand*{\missingcitation}{{\Huge \colorbox{red}{?citation?}}}

\makeatletter
\xpatchcmd{\@setref}{\bfseries}{\missingreference}{}{}
\def\@citex[#1]#2{\leavevmode
    \let\@citea\@empty
    \@cite{\@for\@citeb:=#2\do
        {\@citea\def\@citea{,\penalty\@m\ }%
            \edef\@citeb{\expandafter\@firstofone\@citeb\@empty}%
            \if@filesw\immediate\write\@auxout{\string\citation{\@citeb}}\fi
            \@ifundefined{b@\@citeb}{\hbox{\reset@font\missingcitation}%
                \G@refundefinedtrue
                \@latex@warning
                {Citation `\@citeb' on page \thepage \space undefined}}%
            {\@cite@ofmt{\csname b@\@citeb\endcsname}}}}{#1}}
\makeatother

\newcommand{\mm}[0]{\abr{llm}}

\newcommand{\gem}[1]{\mbox{\textsc{gem}}}
\newcommand{\abr}[1]{\textsc{#1}}

\renewenvironment{quote}
{\list{}{\rightmargin\leftmargin}%
    \item\relax\small\ignorespaces}
{\unskip\unskip\endlist}

\newcommand{\hidetext}[1]{}
\newcommand{\ignore}[1]{}

\ifcomment
    \newcommand{\pinaforecomment}[3]{\colorbox{#1}{\parbox{.8\linewidth}{#2: #3}}}

    \newcommand{\prtodo}[1]{\pinaforecomment{lightblue}{pr}{#1}}
    \newcommand{\prtodoi}[1]{\pinaforecomment{lightblue}{pr}{#1}}
\else
    \newcommand{\pinaforecomment}[3]{}
    \newcommand{\prtodo}[1]{}
    \newcommand{\prtodoi}[1]{}
\fi

\newcommand{\smallurl}[1]{ \begin{tiny}\url{#1}\end{tiny}}

\definecolor{lightblue}{HTML}{3cc7ea}
\definecolor{CUgold}{HTML}{CFB87C}
\definecolor{grey}{rgb}{0.95,0.95,0.95}
\definecolor{ceil}{rgb}{0.57, 0.63, 0.81}
\definecolor{UMDred}{HTML}{ed1c24}
\definecolor{UMDyellow}{HTML}{ffc20e}


%% file: pages/00-abstract.tex
 \begin{abstract}
Event annotation is important for identifying, monitoring, and understanding sociological trends.
Although expert annotators set the gold standard, they are expensive and inefficient.
%
%
%
While state-of-the-art \abr{nlp} models are an attractive alternative, they are often evaluated on standalone subtasks rather than entire workflows. 
Thus, we evaluate a holistic workflow that summarizes news with event coreference resolution and argument extraction in three modes: \abr{ai}-only, \abr{ai} assistance, and human only.
Although \abr{ai}'s recall is seven times higher than the \textsc{tf\nobreakdash-idf} baseline at coreference resolution, it is far from replacing experts.
However, experts \emph{adopt} \abr{ai}-extracted arguments $60\%$ of the time, reducing extraction time by $25\%$. Our code and data are in \url{https://github.com/Obertura777/gtd-data}.

\end{abstract}
%

%

%% file: pages/10-intro.tex

\section{Making Event Annotation Realistic}
\label{sec:intro}

Quality data is crucial for informed decisions~\cite{akella-etal-2025-codegenwrangler, zhang-etal-2025-webquality}. 
Bloomberg\footnote{\url{https://www.bloomberg.com/}} and LSEG\footnote{\url{https://www.lseg.com/}}, for example, gather top-quality data from trusted sources to monitor market trends, while companies like Scale~AI\footnote{\url{https://scale.com/}} collect, curate, and annotate data for \abr{ai} models.

An important but complex step in data processing is cross-document coreference resolution~\cite{grosz-sidner-1986-attention}: finding unique incidents from duplicate and conflicting documents. 
While humans excel at this step, machines handle larger inputs and improve efficiency and scalability~\cite{King_Lowe_2003, 6284096}. 
The Uppsala Conflict Data Project, one of the earliest event dataset programs, covers fewer than \num{300000} events with human labor. 
In comparison, automated processes like \abr{gdelt}~\cite{leetaru2013gdelt} collect trillions of events despite starting later. Such datasets enable organizations to monitor trends and detect anomalies, driving lasting impact in their communities. 

Automating event annotation presents significant challenges. 
Models struggle with text extraction~\cite{Schrodt2013}: empirical evaluations yield low confidence and inconsistent results~\cite{obrien}.
Annotating original sources requires expertise in synthesizing information from multiple documents~\cite{gao-etal-2024-harvesting, li2024improvingtenorlabelingreevaluating}, often involving steps beyond labeling. 
Realistic event datasets in social science research typically combine manual efforts~\cite{elliott_code, doi:10.1177/1077800414532435} with quantitative methods~\cite{PROBIERZ20223449, li2025largelanguagemodelsstruggle} to ensure consistency and data quality.


While large language models (\mm{}s) are more powerful than methods used in real-world applications~\cite{chen-etal-2023-cross, zhao-etal-2025-beyond}, existing pipelines such as \abr{acled} and \abr{icews} do not use them. First, data characteristics differ. 
Unlike deployed workflows that collect millions of documents periodically~\cite{5a745627-8fd8-35a0-afd0-5e17f9e148db, fff7b163-d724-3ebd-8562-84151d5aa987, garcia-duran-etal-2018-learning,leetaru2013gdelt}, \abr{nlp} datasets like \abr{ace} and \abr{ecb} have fixed sizes, few distinct components, and little lexical diversity~\cite{zhukova-etal-2022-towards, doddington-etal-2004-automatic}. 
They also use secondary sources~\cite{ahmed-etal-2024-generating}, under-trained workers~\cite{sharif-etal-2024-explicit}, and document-level scope~\cite{ebner-etal-2020-multi, li2021documentleveleventargumentextraction}.
Lacking source diversity introduces overfitting: zero-shot \abr{gpt}-4o-mini achieves $0.83$ $F_1$ on the six event types of the Gun Violence Corpus~\cite{vossen-etal-2018-dont} and $63$\% accuracy on \num{710} examples in \abr{maven-arg}~\cite{wang-etal-2024-maven} in our test, showing these tasks are becoming less challenging for fast-evolving models.


Second, specialized models are inaccessible. 
They are scheme-dependent and quickly become obsolete as the field changes~\cite{chen-etal-2023-cross, li2021documentleveleventargumentextraction}. 
Reproducing their results is also difficult because open-source models often omit components, inflate results by overfitting, and generalize poorly~\cite{gao-etal-2024-harvesting, liu2024singleeventextractionefficientdocumentlevel, bugert2021generalizingcrossdocumenteventcoreference, cattan-etal-2021-realistic}. 
A general-purpose model is easier to adapt for organizations where these realistic annotations take place.

Third, \abr{nlp} research on event annotation focuses on dedicated tasks but not holistic systems. 
While focusing on one task (e.g., event extraction) allows dedicated contributions, it undermines downstream applicability for social science annotations, which chain individual tasks into a complete pipeline. 
Without reliably combining components like filtering, retrieval, and coreference resolution, end-to-end applicability remains low: actively-maintained pipelines still use statistical methods and rely on trained annotators~\cite{ACLEDMethodology2017, 9526c443-9ec7-3ca1-855c-73628231d44f}.


To address these challenges, we present a case study on \globalterrorism~(\gtd{})\footnote{\url{https://www.start.umd.edu/data-tools/GTD}}, an open-source event database used to systematically study terrorism events. 
This dataset reflects current annotation practices in social science research: it is large, recent, noisy, first-hand, and expert-annotated. 
We detail the operational background of this workflow in Section~\ref{sec:bg}.

We evaluate this pipeline by comparing a manual workflow with one involving \mm{}s.
We test \mm{} capabilities in coreference resolution (Section~\ref{sec:curation}), then measure their impact on event argument extraction (Section~\ref{sec:annotation}). 
Using operational metrics like variable extraction frequency, $F_1$ against expert annotations, and reduction in annotation time, we show that while \mm{}s do not reach expert quality, even a small model effectively assists trained annotators by reducing annotation time. 

Finally, we outline how to practically achieve \mm{} integrations by summarizing the common errors \mm{}s still produce (Section~\ref{sec:err}), highlighting mitigation strategies for live deployment (Section~\ref{sec:mit}), and contextualizing our approach within related work (Section~\ref{sec:rel}).

%% file: pages/20-background.tex
\section{A Realistic Workflow}
\label{sec:bg}
From a list of unorganized documents, our goal is to form a set of incidents with annotated variables and references to relevant documents. 
To achieve this, annotators start with a massive but noisy pool of documents gathered through automated ingestion, which removes easily identifiable duplicates and calculates document similarities before manual work begins.

Once the automated process reduces the document count, experts engage in the first manual step: \cdcr{}.
Experts identify unique events from all documents and construct \incidentset{1}s, sets of documents about single events like violent protests. 

This process is akin to cross-document coreference resolution~\cite{bagga-baldwin-1998-entity-based}. Annotators must review all documents to consolidate scattered details, as single documents often contain only partial, complementary, or even conflicting information. 
To help find similar documents, the automated system generates relevance scores via \textsc{tf\nobreakdash-idf} and nearest-neighbor search, allowing annotators to view similar texts (Figure~\ref{fig:phase_1_review_article}). 

Because we need consistent \incidentset{1}s to fairly evaluate downstream annotations, we first isolate this curation step. We evaluate candidate \incidentset{1}s using precision, recall, and $F_1$ against expert-created baselines, giving a clear metric for model helpfulness in clustering. 

Once \incidentset{1}s are established, the second step is event argument extraction, which \gtd{} calls \cdeae{}. 
Here, experts synthesize the curated text to extract domain variables defined in a formal codebook. 
These variables range from free-text descriptions (e.g., target of ``police vehicle'' in a terrorism event in Pakistan) and enumerated categories (e.g., attack types such as bombing, armed assault, and facility/infrastructure attack) to numerical values like casualty counts. 

To measure the impact of \mm{} assistance on this step, we test three experimental conditions: humans annotating alone, humans vetting \mm{}-generated outputs, and \mm{} outputs alone, following conventions in human\nobreakdash-\abr{ai} collaboration~\cite{Sheridan1978HumanAC, 10.3389/frobt.2022.782134}. We assess practical utility by measuring total annotation time with and without \mm{}-coded variables, tracking how often humans accept \mm{} suggestions, and calculating overall annotation accuracy.

%% file: pages/30-incident_set_curation.tex
\section{Methods for \cdcr{}}
\label{sec:curation}
\input{tables/curation}
To evaluate the effectiveness of \mm{}s in \cdcr{}, experts manually review \num{500} documents from February 2022\footnote{We deliberately sampled from this period because the onset of the Russo-Ukrainian war generates many terrorism-related events not previously accounted for.} to identify events and create gold-standard \incidentset{1}s. 
Given the same documents, we then use three methods to generate \incidentset{1}s that contain these documents. We compare them against this gold standard.

\textbf{\cls{}} (Pairwise Classification): \abr{gpt}-4o-mini\footnote{We choose it for its availability and cost-effectiveness as larger models do not necessarily yield significantly better results~(Appendix~\ref{app:justification}).} classifies document pairs. It compares a candidate document not yet in the output \incidentset{1}s against a reference document to predict whether both texts describe the exact same event. This method also includes an optional pre-processing step (\textsc{seg}) that uses the \mm{} to segment long, multi-topic digest documents into single-event text blocks before classification. 

\textbf{\emb{}} (Embedding Similarity): We group related documents based on their overarching semantic proximity rather than relying on exact keyword matches. To achieve this, we calculate the pairwise cosine similarity of vector embeddings between a new document and the existing members of a cluster. A document joins the cluster if its similarity score to any member exceeds $0.859$, a threshold established by a grid search over a separate, held-out validation set that optimize for the higest $F_1$ score\footnote{We show the algorithm in Appendix~\ref{sec:alg}.}.


\textbf{\abr{K\nobreakdash-LLMmeans}}: This method uses \abr{k-means} with \mm{} summarization to iteratively update cluster centroids and assign documents~\cite{diazrodriguez2025kllmmeanssummariescentroidsinterpretable}. We show our configuration and a summary example in Appendix~\ref{app:kllmmeansconfig} and \ref{app:sum}. 


\subsection{Improving \cdcr{}}
\textbf{\emb{} captures semantic relationships.} 
Mapping text into dense vector spaces, \emb{} captures the underlying meaning rather than relying on word matches. This allows it to group documents effectively and achieves $0.89$ precision, a significant improvement over the traditional \textsc{tf\nobreakdash-idf} baseline (i.e., scores generated by the existing automated system). \emb{} reduces false negatives by $82$\% and successfully identifies $51$\% of all relevant documents.

\textbf{Segmentation helps.} Real-world documents are messy; news digests contain multiple unrelated events. Dividing them into discrete event sections before classification improves \cdcr{}. Segmentation reduces ambiguity and prevents the \mm{} from misidentifying events, yielding the best recall and $F_1$ score (Table~\ref{tab:curation}). 

\textbf{\abr{K\nobreakdash-LLMmeans} has higher accuracy and lower cost than \cls{}.} 
Although segmentation paired with pairwise classification increases recall, binary predictions are not cost-effective for large corpora because of quadratic complexity. 
However, by combining high-quality embeddings with \mm{}-generated summaries, \abr{K\nobreakdash-LLMmeans} achieves a higher recall and $F_1$ score than \cls{} at less than one-third of the cost.

%

%% file: tables/curation.tex
\begin{table}
\centering
\resizebox{\columnwidth}{!}{
\begin{tabular}{lrrrr}
 \toprule
 & Precision & Recall & $F_1$ & Same Sets \\
 \midrule
  \abr{tf-idf}&0.19&0.09&0.10&12\\
 \midrule
 \multicolumn{5}{l}{Pairwise} \\
 \textsc{embedding}&\textbf{0.89}&0.51&0.59&66\\
 \textsc{llm-cls}&0.36&0.35&0.35&105\\
 ~~~ \textsc{+seg}&0.65&\textbf{0.66}&\textbf{0.63}&\textbf{203}\\
 \midrule
 \multicolumn{5}{l}{\abr{K\nobreakdash-LLMmeans}} \\
 \textsc{text-embedding-3-small}&0.36&0.45&0.38&85\\
 \textsc{gemini-embedding-001}&0.32&0.36&0.32&60\\
 \textsc{DistilBERT}&0.12&0.10&0.09&7\\
 \textsc{ModernBERT}&0.20&0.20&0.16&15\\
 \bottomrule
\end{tabular}
}
\caption{In \cdcr{}, \emb{} has the highest precision of creating \incidentset{1}s. \textsc{llm-cls+seg} shows superior recall and a higher overall $F_1$ score. Additionally, compared to $371$ annotated\incidentset{1}s, \textsc{llm-cls+seg} generates \incidentset{1}s that align most closely with expert annotations. \textsc{K-LLMmeans} is more cost-effective than \textsc{llm-cls} with better recall and $F_1$.}
\label{tab:curation}
\end{table}

%% file: pages/40-annotation.tex
\section{\cdeae{} with \mm{}s}
\label{sec:annotation}
\input{figures/annotation}
A holistic workflow includes multiple \abr{nlp} tasks. 
The quality of \cdcr{} determines the success of \cdeae{}. 
In \cdcr{}, we prioritize recall over precision because it ensures annotators have coverage of relevant documents in this stage. 
While \emb{} has the highest precision, \seg{} has the highest recall and generates many \incidentset{1}s identical to expert annotations ($203$ out of $371$, Table~\ref{tab:curation}).

In \cdeae{}, we assess how flawed \incidentset{1}s affect annotation by replacing some expert-curated events with the most similar \mm{}-generated ones. 
This creates three distinct types of \incidentset{1}s (Figure~\ref{fig:annotation_time}): \textsc{manual} (human-created), \textsc{llm} (\mm{}-generated), and \textsc{consensus}. 
The \textsc{consensus} sets, where experts and \textsc{llm-cls+seg} concur, allow us to compare the pipeline independent of the \incidentset{1} creation method. 

For \textsc{manual}, \textsc{llm}, and \textsc{consensus} respectively, we compare the variables coded by experts operating in two conditions: a hybrid setting (with access to \mm{} suggestions) and a manual setting (without \mm{} assistance). Experts code nine variables for $212$ \incidentset{1}s.

Because humans show ``automation bias'' by over-relying on \mm{}-coded variables~\cite{an2023sodapopopenendeddiscoverysocial}, we compare inter-human agreement among three teams against the human-\mm{} agreement. 

Traditional exact-match metrics are overly strict~\cite{kocmi-etal-2021-ship, chen-etal-2020-mocha}, so we use three evaluation methods:

\textbf{Normalized Match (\textsc{nm}):} Checks if two variables are identical after stripping punctuation, converting to lowercase, and normalizing whitespace.

\textbf{\textsc{bert} Match (\textsc{bem}):} Measures embedding similarity~\cite{bulian-etal-2022-tomayto}.

\textbf{\abr{pedants}:} Uses $F_1$ score and \textsc{tf\nobreakdash-idf} to measure the variable match~\cite{Li2024PEDANTSCB}.

\subsection{\mm{}s Help Experts Code Variables}

\input{figures/arg_fre_fig}
\textbf{\mm{}-coded variables drastically reduce annotation time.} 
Experts annotate fastest when their judgments align with the \seg{} clusters (Figure~\ref{fig:annotation_time}). 
Conversely, coding event variables from purely human-created \incidentset{1}s takes the longest. 
Despite errors in \mm{}-coded variables, simply providing these suggestions to annotators reduces annotation time by $25\%$.


\textbf{Annotators use \mm{}-coded variables two-thirds of the time.} 
Presented with options, experts rely heavily on the \mm{}. They choose the suggestion for the \texttt{Country} variable $92$\% of the time. 
The higher agreement in the hybrid setting across all \incidentset{1} types shows that \mm{}-coded variables aid annotation (Figure~\ref{fig:subset}).

\textbf{Experts agree with \mm{}-coded variables over 55\% of the time,} even without seeing them (Figure~\ref{fig:avg}). 
Because the human-\mm{} agreement nearly matches human-human agreement rate, the \mm{}-coded variables provide a near-human level utility.
\input{figures/subset}

\textbf{Model size does not strongly correlate with accuracy.} 
We test open- and closed-source \mm{}s, expecting larger models to dominate, but none are significantly better. 
Notably, \textsc{mistral 8x7b} shows low accuracy for numerical variables (\texttt{kills} and \texttt{wounds}). 
We advise researchers to benchmark models against their specific codebook if the target variables include non-textual ones.

\textbf{Models hallucinate when data are missing.} 
We calculate precision and recall for scenarios where a variable is missing from the source incident. 
If no casualty count is mentioned, the correct annotation is "Not Available" (\abr{NA}). 
High precision means when the model reports a variable as "\abr{NA}", it is rarely wrong. 
High recall means the model correctly identifies most of the situations where the information is truly not available instead of hallucinating a false value. 
For most models tested, precision hovers around $0.5$ and recall sits around $0.25$, showing strong hallucination. 
The outlier is \textsc{mixtral-8x7b}: showing high precision ($0.62$) but abysmal recall ($0.03$), meaning it almost always hallucinate an answer (Table~\ref{tab:na_results}). 

\section{Error Analysis}
\label{sec:err}
In \cdcr{}, \mm{}-generated \incidentset{1}s correlate poorly with expert judgment, and even the best model falls short of expert accuracy. We describe some error types below and show examples in Appendix~\ref{app:err_exs}.

\textbf{Under-specified Instructions:} Unclear prompts lead to imprecise and missing details. 
For example, experts frequently select the suggested \texttt{country} variable but rarely \texttt{location} because \texttt{country} is rigidly defined while \texttt{location} is vague, ranging from broad regions to specific sites. 

\textbf{Source Document Ambiguity and Temporal Conflict:} Dense cross-subtopic links ($2.95$ documents per event on average) mean digest documents summarize multiple events, misleading \abr{gpt}-4o-mini. 
A digest document, common in our data, can also summarize multiple events, which misleads \abr{gpt}-4o-mini into annotating the wrong event. 
Additionally, temporal components also appear in the news cycle.

\textbf{Interpretive Subjectivity:} Human judgments affect \cdeae{}. \mm{}s lack access to the \gtd{} codebook conventions that disambiguate overlapping categories.

\subsection{Mitigation Strategies}
\label{sec:mit}
Refined prompts reduce these errors. 
Providing \mm{}s with contexts, additional instructions and task information makes them more robust. 

%% file: figures/annotation.tex

\begin{figure}
    \centering
    \includegraphics[width=1.0\linewidth]{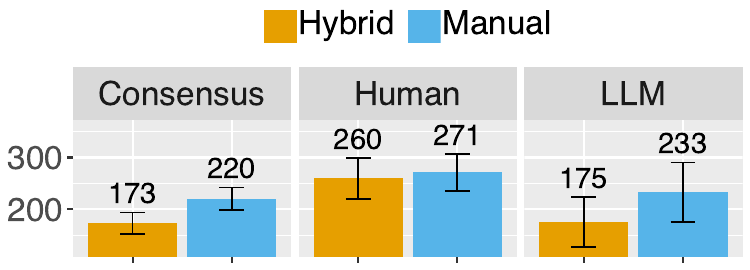} 
    \caption{Experts need less time (in seconds) in \cdeae{} when they have access to an \mm{} (i.e., hybrid) under all three conditions. Consensus are the event sets that human and \textsc{ai} agree, in which the time taken is the lowest. Annotators take less time in the \textsc{llm}-generated event sets because many are invalid. When experts and \textsc{llm-cls+seg} agree on event sets, they take the least amount of time with lower deviations.}
    \label{fig:annotation_time}
\end{figure}


%% file: figures/arg_fre_fig.tex
\begin{figure}
    \centering
    \includegraphics[width=1.0\linewidth]{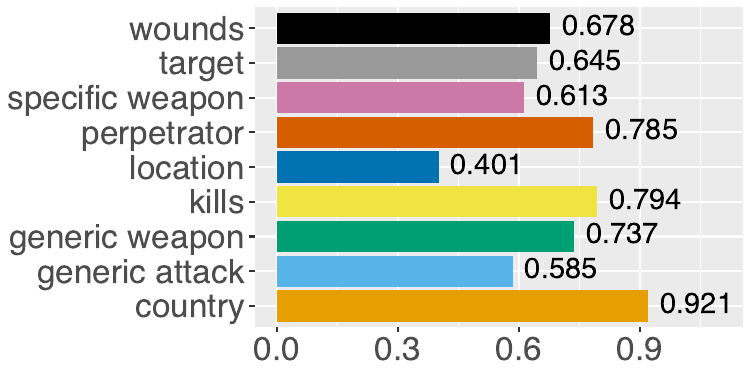} 
    \caption{How often expert annotators use \mm{}-coded variables when given the chance in \cdeae{}. While the selection frequency varies, even the least selected variable, \texttt{location}, is useful 40\% of the time.}
    \label{fig:arg_fre}
\end{figure}

%% file: figures/subset.tex
\begin{figure*}[t]
    \centering
    \includegraphics[width=1.0\linewidth]{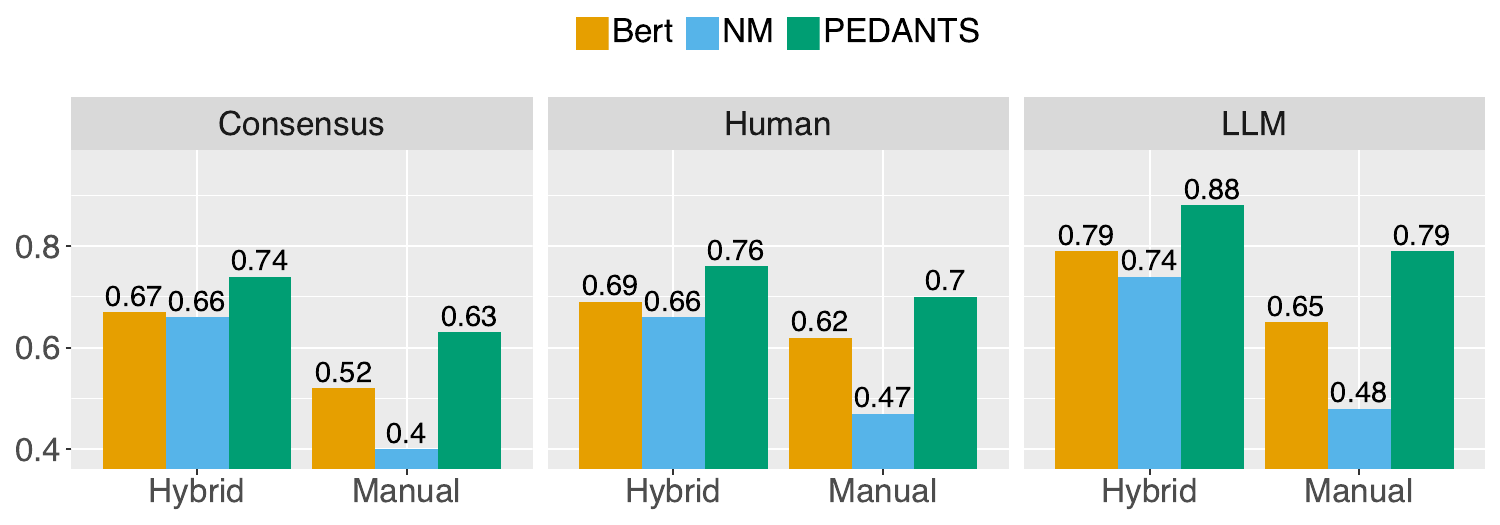} 
    \caption{Models help annotate the variables. Annotators show higher agreement in the hybrid setting, where \mm{}-coded variables are available. The variables prove particularly beneficial in \mm{}-generated \incidentset{1}s, which often contain misinformation.}
    \label{fig:subset}
\end{figure*}

%% file: pages/50-related_work.tex
\section{Related Work}
\label{sec:rel}
\noindent \textbf{Automated Event Data Annotation:} Early efforts to scale event data collection incorporate traditional \abr{nlp} techniques, such as statistical retrieval systems, to efficiently filter content and automate extractions~\cite{D’Orazio_Landis_Palmer_Schrodt_2014, 6284096, Boschee2013}. As event attributes become more granular and complex, automated extraction accuracy rapidly decreases~\cite{Jenkins_Maher_2016}. Automated datasets are notorious for containing duplicates, erroneously including non-events, and generating false positives, largely stemming from geo-localization errors and rigid syntactic parsing~\cite{Hammond_Weidmann_2014, Miller_2022, Raleigh_Linke_Kishi_2023}. Consequently, accurate, high-fidelity event databases still require human labor. Our work bridges this gap by positioning \mm{}s not as fully autonomous replacements, but as expert assistants. 

\noindent \textbf{\mm{}s for Event Tasks:} Prior work shows \mm{}s are cost-effective document-level annotators for event~\cite{Chen_Qin_Jiang_Choi_2024} and argument extraction~\cite{shuang-etal-2024-thinking, zhang-etal-2024-ultra, zhu2023chatgptreproducehumangeneratedlabels, wang-etal-2021-want-reduce}. 
However, \mm{}s struggle with financial data~\cite{tseng2024expertlevellanguagemodelsexpertlevel} and word semantics~\cite{yadav2024automatingtextannotationcase}. 
While \citet{zhao-etal-2023-cross} show that \abr{gpt}-4 achieves higher accuracy than under-trained crowd workers, we apply \mm{}s to a more complex workflow where annotators must first identify relevant events before extracting arguments from them. 

%% file: pages/60-conclusion.tex
\section{Conclusion}
\label{sec:con}
Realistic event annotation remains a complex, resource-intensive process that depends on trained human labor. We present a case study to integrate \mm{}s to alleviate human effort in a holistic pipeline. 
In \cdcr{}, we show that \mm{}-based segmentation and clustering methods achieve higher precision and recall than the \textsc{tf\nobreakdash-idf} baseline, hence helping annotators to find similar documents, despite not able to automate the workflow.
During \cdeae{}, we find that providing \mm{}-coded variables to experts reduce annotation time by $25$\%. 
Future work can explore fine-tuning with task rewards or domain-specific datasets to further encourage \mm{} adoption. Task-agnostic prompt-tuning and few-shot learning offer paths to increase model accuracy~\cite{khattab2024dspy, wang-etal-2022-towards-unified, gao-etal-2021-making}.


%% file: pages/70-limitations.tex
\section{Limitations}
\label{sec:limitations}
\abr{nlp} techniques have not yet reached human-level accuracy in document classification. In \cdcr{}, the computational cost of pairwise similarity increases quadratically with the linear growth in the number of documents, rendering \mm{}\nobreakdash-based methods inefficient for large document sets (except \abr{K\nobreakdash-LLMmeans}). Although it is impractical to fully replace \textsc{tf\nobreakdash-idf}, \mm{}-based methods are useful in finding semantically similar documents when \textsc{tf\nobreakdash-idf} are the same. To alleviate computational efforts, instead of computing pairwise embedding similarity for every document, we only compute pairwise embedding similarity for documents above the \textsc{tf\nobreakdash-idf} threshold. For \cdeae{}, annotation requires greater granularity to meet criteria outlined in specified guidelines. These steps are essential for implementing \mm{}\nobreakdash-assisted large-scale event data collection workflows. Additionally, we have not tested \mm{}s in cross-document event coding, which is a useful step for mitigating \cdeae{} errors.

\section*{Acknowledgments}
We thank researchers from \gtd{}. Specifically, our thank goes to Jacob Scott Loewner, Margaret A. Hayden, Oleksiy Krylyuk, and Tyler Yates for data annotation and its discussion. We thank Brian Wingenroth and Dr. Amy Pate for their insights on the \gtd{} workflow. This research was funded in part by the Department of Defense under award no. HQ003421F0481. Any opinions, findings, and conclusions or recommendations expressed in this report are those of the authors and do not necessarily reflect the views of the Department of Defense.

%% file: pages/80-appendix.tex
\appendix

\section{Appendix}
\label{sec:appendix}

\subsection{Extended Related Work}

\noindent \textbf{Cross-Document Event Tasks:} Cross-document \cdcr{} and \cdeae{} are more challenging than document-level ones because they require identifying events and selecting documents from a noisy group~\cite{bornstein-etal-2020-corefi, Cattan2020StreamliningCC, ji-etal-2009-cross, barhom-etal-2019-revisiting}. 
Prior work uses discourse structure~\cite{chen-etal-2023-cross}, Transformer encoders~\cite{gao-etal-2024-harvesting}, negative examples~\cite{cattan-etal-2021-cross-document}, and retrievers~\cite{eirew-etal-2022-cross}. 
The work by \citet{zhao-etal-2023-cross} most closely resembles a realistic setting; they collected over \num{1500000} COVID-19 documents but evaluated on only 100. 
In contrast, we use a fivefold larger test set and both the \cdcr{} and \cdeae{} tasks.

\noindent \textbf{Text Segmentation and Clustering:} Because real-world news articles often digest multiple events, text segmentation is crucial for improving pipeline recall. Historically, algorithms like TextTiling~\cite{10.5555/972684.972687} and TopicTiling~\cite{riedl12_acl} manage texts with highly distinct sections. Later neural approaches treated segmentation as a supervised task, using bi-directional \abr{lstm}s and \abr{bert}-based models trained on dataset like Wikipedia or legal documents~\cite{koshorek-etal-2018-text, Aumiller_2021}. However, documents containing closely related, overlapping events like weekly regional conflict roundups post greater segmentation challenges than articles with distinct topical boundaries. Our use of \mm{}-based segmentation with \abr{K\nobreakdash-LLMmeans} address the limitations of inefficient clustering and semantic understanding.

\input{figures/var}
\subsection{Automated Data Processing at \gtd{}}
\label{app:source}


A retrieval model collects news articles from trusted sources like LexisNexis and BBC Monitoring. String filters identify documents containing potential attack information, focusing on keywords such as \emph{assault}, \emph{hostage}, and \emph{rebel}. Then, an algorithm removes duplicates and irrelevant documents from this pool. A Support Vector Machine uses \textsc{tf\nobreakdash-idf} to flag highly relevant documents for manual review. In \cdcr{}, our baseline, $k$-\textsc{nn} search, displays documents within a 1.35 radius using a \textsc{tf\nobreakdash-idf} vectorizer.

\subsection{Definition of An Event and Inclusion Criteria}
\label{sec:def}
The \gtd{} team defines a terrorism event as the threatened or actual use of illegal force and violence by a non-state actor to attain a political, economic, religious, or social goal through fear, coercion, or intimidation. Specifically, a document must have all of the following attributes to be included in the database:
\begin{enumerate*}
    \item \textbf{The event must be intentional} - the result of a conscious calculation on the part of a perpetrator.
    \item \textbf{The event must involve violence} - against either property or people.
    \item \textbf{The perpetrators of the events must be sub-national actors.} State-level  is excluded from the database.
\end{enumerate*}

In addition, the document must also meet at least two of the following criteria:
\begin{enumerate*}
    \item \textbf{The act must be aimed at attaining a political, economic, religious, or social goal.} In terms of economic goals, the exclusive pursuit of profit does not satisfy this criterion. It must involve the pursuit of more profound, systemic economic change.
    \item \textbf{There must be evidence of an intention to coerce, intimidate, or convey some other message to a larger audience (or audiences) than the immediate victims.}  It is the act taken as a totality that is considered, irrespective if every individual involved in carrying out the act was aware of this intention. As long as any of the planners or decision-makers behind the attack intended to coerce, intimidate or publicize, the intentionality criterion is met.
    \item \textbf{The action must be outside the context of legitimate warfare activities.} That is, the act must be outside the parameters permitted by international humanitarian law, insofar as it targets non-combatants.
\end{enumerate*}

\subsection{Variable Schema}
\label{app:vcs}
Here, we describe the variables in the annotation process, specified by \gtd{}.
\begin{enumerate}
    \item \textbf{Country:} the country in which the  event occurred.
    \item \textbf{Location:} the most specific location (e.g., village name) in which the  event occurred.
    \item \textbf{Target:} the targeted group of the  event.
    \item \textbf{Perpetrator:} the group carrying out the  event.
    \item \textbf{Generic Attack Type:} One or more of \emph{Facility/Infrastructure Attack}, \emph{Armed Assault}, \emph{Assassination}, \emph{Bombing/Explosion}, \emph{Hostage Taking (Kidnapping)}, and \emph{NA}.
    \item \textbf{Generic Weapon:} One or more of \emph{Explosives}, \emph{Firearms}, \emph{Incendiary}, \emph{Sabotage Equipment}, \emph{Melee}, \emph{Vehicle}, and \emph{NA}.
    \item \textbf{Specific Weapon:} A detailed description of \emph{Generic Weapon}.
    \item \textbf{Kills:} Number of people killed during the  event.
    \item \textbf{Wounds:} Number of people injured during the  event.
\end{enumerate}

\subsection{Attack Type Distribution}

\input{tables/event_type_distr}

\subsection{Generic Weapon Type Distribution}

\input{tables/weapon_type_distr}

\subsection{An Illustration of Annotation Difficulty}
In \cdcr{}, one document describes an event ``\emph{on Sunday, where a man attack a vehicle in Barangay Palampas,}'' while another specifies ``\emph{the incident occurred in Barangay Palampas, San Carlos City, Negros Occidental, on February 20,}'' providing precise details about the event's date. 

In \cdeae{}, one document describes an event involving ``\emph{approximately 20 people, some with axes, who caused injuries}'' in ``\emph{Houston, British Columbia,}'' while another document states ``\emph{far-left anti-pipeline extremists}'' attacked the ``\emph{Morice River drill pad site off the Marten Forest Service Road.}''

\subsection{Annotator Background and Training}
Only full-time \gtd{} researchers are involved in \cdcr{}, since it is regarded as a more complex and difficult process than \cdeae{}. In \cdeae{}, student interns and part-time research assistants under the supervision of full-time researchers record detailed information about each attack. Everyone on the \gtd{} team has a background in terrorism/political violence. All annotators on this project hold at least an MA. 

One training presentation focuses on the interface---what are the various features of the interface, how to navigate the document window, similar documents, how to identify existing events, etc. The second presentation focuses on the inclusion criteria and an overview of the codebook. New annotators are not expected to know all of the details and nuances for each variable, but they should be able to grasp the top-level variables as well as understand what information to look for in documents that might be relevant to each coding domain (locations, weapons/tactics, casualties, perpetrators, targets, general). These presentations are split to avoid overwhelming trainees with information, but together they probably comprise about 8 hours.

Then, the trainees are put into a sandbox loaded with real documents from previous months. The supervisor provides feedback on those annotations. Once the supervisor is satisfied with the work, the trainee is taken "out of the sandbox" and allowed to triage with the rest of the team. There is not a fixed number of hours that a person stays inside the sandbox as people understand things at different speeds but the process can take weeks. Trainees go through these documents as they would in the real data, identifying events and collecting documents with the relevant information. The supervisor gets files with their progress (events created, documents added to events, as well as documents discarded) and evaluates it, then reviews the feedback with the trainee. There are multiple rounds of work and feedback in the sandbox until the trainee is proficient to begin triaging realistically. The length of this process depends on each new annotator but generally takes a few weeks. New annotators are also spending time doing other tasks (e.g. coding, supervising interns), a time estimate for this stage of the training is 25-40 hours.

In addition, these annotators will often have previous experience with \gtd{} (e.g. coding experience as an intern or hourly), or they will be gaining experience in their role on one of the coding domains, through which they will be getting in-depth knowledge on a section of the codebook, which should reinforce their annotation expertise.

\subsection{Algorithm for Finding the Best Embedding Threshold}
\label{sec:alg}

\begin{algorithm}
\caption{Embedding Algorithm}\label{alg:embedding}
\begin{algorithmic}[1]
    \State \textbf{Input:} list of documents
    \State \textbf{Output:} best precision, recall, $F_1$
    \State best precision, recall, $F_1$ $\gets 0, 0, 0$
    
    \For{$i = 1$ to steps}
        \State threshold $\gets$ min $+ \frac{\text{(max - min)} \cdot i}{\text{steps}}$
        
        \For{each document1 in all documents}
            \State similars $\gets \emptyset$
            \For{each document2 in all documents}
                \State cal\_sim(document1, document2)
                \If{$\text{similarity} \geq \text{threshold}$}
                    \State similars.add(document2)
                \EndIf
            \EndFor
        \If{$F_1(\text{ref}, \text{similars}) > \text{best $F_1$}$}
            \State update best $F_1$
        \EndIf
        \EndFor
    \EndFor
    \State \Return best (precision, recall, $F_1$)
\end{algorithmic}
\end{algorithm}

\subsection{Over- and Under-generation of \abr{llm-cls}}
\label{sec:lap}
Using an \mm{} to generate \incidentset{1}s candidates would almost always create different numbers of\incidentset{1}s compared to \textsc{tf\nobreakdash-idf}, making direct comparison difficult. To address potential over- or under-generation, we formulate this as a linear assignment problem, optimizing the average $F_1$ score between the results and the human-coded reference set.

The linear assignment problem is an optimization problem. The objective is to assign a gold \incidentset{1} to a generated \incidentset{1} in such a way that the overall cost is minimized. $\mathcal{C}_{i, j}$ represents the cost of matching \incidentset{1} $i$ in the gold set with \incidentset{1} $j$ in the prediction set. Formally, the optimal assignment has cost
\[
    \sum_{i} \sum_{j} \mathcal{C}_{i, j},
\]
where 
\[
    \mathcal{C}_{i, j} = -F_1{(i,j)}
\]

\subsection{Variable Coding Error Examples}
\label{app:err_exs}
\textbf{Interpretive Subjectivity:} One document mentions the administrative area ``\emph{Pale, Sagaing}'' and the more specific ``\emph{Einmahti village}'' in Myanmar, the model selects the broader region, whereas experts prefer the latter. 

\textbf{Temporal Conflict:} One initial breaking report states ``\emph{five people injured,}'' while an updated report later revises the count to ``\emph{at least eight people wounded.}'' Under-trained annotators and \mm{}s alike struggle to resolve these timelines to find the accurate information. 

\textbf{Interpretive Subjectivity:} An attack on non-human entities is 
\texttt{facility/infrastructure} if buildings are the primary target; otherwise, they use \texttt{armed assault}. 
This dividing line is often subjective.
Assassinations via explosives are classified as \texttt{assassination} in the \gtd{} codebook, not \texttt{bombing/explosion}. 
Lacking this knowledge, \mm{} codes those variables more frequently as \texttt{bombing} and \texttt{armed assault}.

When we include variable definitions in Appendix~\ref{app:vcs} into the prompt, semantic accuracy (measured by \textsc{bem} and \textsc{pedant}) increases for \texttt{specific weapon}, \texttt{location}, \texttt{target}, \texttt{perpetrator}, \texttt{wounds}, and \texttt{generic attack} with statistical significance ($p<0.05$).

\subsection{Human-\mm{} Agreement}
We ask three groups of annotators to annotate a shared portion of the \incidentset{1}s and check the inter-annotator agreement on variables (human-human). We also calculate the agreement between annotators operating in the hybrid and the manual setting (human-\mm{}). 
Figure~\ref{fig:avg} displays the results: human-human and human-\mm{} agreements do not differ significantly.
Figure~\ref{fig:var} shows human-\mm{} agreement by variable types. Annotators show 0.89 agreement with \emph{Country}, a variable with a high degree of specificity. In contrast, annotators agree with \emph{Location} infrequently, suggesting less utility of variables with a lower degree of specificity. 
We also investigate if human-\mm{} agreement differ by providers and models (Table~\ref{tab:agg_acc_agreement}).

\input{figures/average}
\input{tables/agreement_acc_agg}

\subsection{Model/Prompt Specifications and Justification for Model Selection}
\label{app:justification}
We tested the following \mm{}s: \href{https://huggingface.co/Qwen/Qwen3-Next-80B-A3B-Instruct}{\textsc{Qwen3-Next-80B-A3B-Instruct}}, \textsc{Gemini 2.5 Pro}, \textsc{GPT-4.1 Mini}, \href{https://huggingface.co/mistralai/Mixtral-8x7B-Instruct-v0.1}{\textsc{Mixtral 8x7B Instruct}}, \textsc{Claude Sonnet 4}, \href{https://huggingface.co/meta-llama/Llama-3.3-70B-Instruct}{\textsc{Llama 3.3 70B Instruct}}, \href{https://huggingface.co/mistralai/Mistral-Small-3.2-24B-Instruct-2506}{\textsc{Mistral Small 3.2 24B}}, \textsc{Gemini 2.5 Flash}, and \textsc{GPT-4o-mini}. All models are zero-shot with 0 temperature. However, output might change for future models.

Upon preliminary inspection, no single model has significantly higher accuracy in \cdcr{} and \cdeae{}. The specific model we used is \textsc{gpt-4o-mini-2024-07-18} because it is cost-efficient. It does not require fine-tuning or computational resources. Social science researchers can use well-developed API calls with minimum cost compared to larger and more advanced models. We show the results of all models in Tables~\ref{tab:na_results} and \ref{tab:agg_acc_agreement}.

\subsubsection{Prompt for \cdcr{}}
\begin{quote}
    \emph{Determine whether the following articles describe the same incident:}\\
    \texttt{\{document 1\}}\\
    \texttt{\{document 2\}}
\end{quote}

\subsubsection{Prompt for \abr{seg}}
\begin{quote}
    \emph{The following document describes zero or more incidents. Segment the document based on incidents mentioned and return an array.}\\
    \texttt{\{document\}}
\end{quote}

\subsubsection{Prompt for \abr{K\nobreakdash-LLMmeans}}
\begin{quote}
    \emph{The following is a set of documents describing a single terrorism event. Write a concise summary that represents the cluster.}
\end{quote}

\subsubsection{System Prompt for \cdeae{}}
\begin{quote}
    \emph{You are a trained annotator. Extract the relevant information based on the given question. Respond with 'NA' if the information is not present in the text. ONLY provide the answer, without any additional explanation.}
\end{quote}

\subsubsection{Prompt for \cdeae{}}
\begin{quote}
    \emph{What is \textsc{event variable} in the event?}
\end{quote}

\subsubsection{Prompt for Error Mitigation}
\begin{quote}
    \emph{What is \textsc{event variable}? DEFINITION: \textsc{event variable definition}}
\end{quote}




\subsection{Model Accuracy in Identifying NA Variables}
See Table~\ref{tab:na_results} and~\ref{tab:agg_acc_agreement}.

\input{tables/na_results}

\subsection{Generative AI Use}
We use generative AI for writing and coding. For writing, we only use AI for checking grammatical and syntax mistakes. For example, making sure we use LaTeX macros instead of string literals and correcting typographical errors and subject-verb agreements. We do not use generative AI in creating research ideas and writing the draft. For coding, we use generative AI for assisting in creating figures and retry decorators for API calls.

\subsection{\abr{K\nobreakdash-LLMmeans} Configurations}
\label{app:kllmmeansconfig}
 We maintain similar settings by using \textsc{text-embedding-3-small} from OpenAI and DistilBERT~\cite{sanh2020distilbertdistilledversionbert}. However, because document length exceeds $512$, DistilBERT's maximum input sequence length, we also add ModernBERT~\cite{warner2024smarterbetterfasterlonger}.  We use the same parameters as \abr{K\nobreakdash-LLMmeans}-5: $120$ iterations with five summarization steps ($l = 20$) and all documents as inputs at each step. Overall, the cost of generating $2,355$ \textsc{text-embedding-3-small} embeddings ($500+5*371$) and $5,565$ ($3 * 5 * 371$) summarizations for all three embedding types is \$13.67, much less than directly comparing documents pairwise.

\subsection{Example of \abr{K\nobreakdash-LLMmeans} Centroid Summaries}
\label{app:sum}
\vspace{1em}
\hrule
\vspace{0.5em}

\paragraph{Iteration 20}
\begin{quote}
Over the course of 40 hours, the Baloch Liberation Army (BLA), an armed group advocating for Balochistan's independence, conducted a significant attack on the Pakistani Frontier Corps (FC) headquarters in Panjgur, Balochistan. The BLA successfully occupied the headquarters, outmaneuvering the Pakistani military despite its superior resources, resulting in substantial casualties among Pakistani troops. This major offensive is the BLA's most daring against what they consider the oppressive and exploitative Pakistani forces. The BLA regards the Pakistani government and its Chinese partners as colonial powers exploiting Balochistan's resources, leaving the local population disenfranchised and impoverished. The attack occurs amid heightened tensions over the China-Pakistan Economic Corridor (CPEC) projects in Balochistan, viewed by locals as marginalizing and excluding them from economic benefits. Recent BLA attacks have targeted Chinese interests, signaling the group's opposition to Chinese involvement in the region. Previously, notable attacks attributed to the BLA include targeted assaults on Chinese nationals and projects in Balochistan, Karachi, and Gwadar, underscoring the group's resolve against external exploitation. These operations have prompted demands from China for enhanced security measures for its nationals in Pakistan.
\end{quote}

\vspace{0.5em}\hrule\vspace{0.5em}

\paragraph{Iteration 40}
\begin{quote}
Over the last 40 hours, the Baloch Liberation Army (BLA), a militant group advocating for Balochistan's independence, executed a significant attack on the Pakistani Frontier Corps (FC) headquarters in Panjgur, Balochistan. This attack, which occurred on Wednesday night and continued until Friday evening, resulted in heavy casualties for the Pakistani forces, despite their sophisticated military equipment. This marks the most daring operation by Baloch organizations against what they perceive as Pakistani occupying forces, who allegedly suppress the Baloch community and exploit their region's resources in collaboration with China, particularly through the China-Pakistan Economic Corridor (CPEC) projects. The BLA's Majeed Brigade, known for its committed fighters, has a history of audacious attacks targeting Chinese interests and Pakistani assets, illustrating the growing unrest and resistance within Balochistan against perceived foreign-backed exploitation. Recent years have witnessed several high-profile BLA attacks, including assaults on Chinese engineers and installations, culminating in deteriorating security conditions in the region for both Pakistani and Chinese stakeholders.
\end{quote}

\vspace{0.5em}\hrule\vspace{0.5em}

\paragraph{Iteration 60}
\begin{quote}
In a significant escalation, the Baloch Liberation Army (BLA) attacked and occupied the Pakistani Frontier Corps headquarters in Panjgur, Balochistan, for nearly 40 hours, marking one of the largest and boldest assaults by the group. This attack, which resulted in heavy casualties among Pakistani troops, underscores the deep-seated grievances of Baloch organizations against the Pakistani government, which they view as a colonial occupier exploiting Balochistan's resources in partnership with China. The BLA's Majeed Brigade, known for its audacious attacks, has historically targeted interests shared by Pakistan and China, driven by marginalization and exploitation concerns, particularly related to the China Pakistan Economic Corridor (CPEC). This incident is part of a series of attacks that include high-profile assaults on Chinese nationals and infrastructure, illustrating a broader struggle for autonomy and recognition by Baloch groups who feel alienated in their region.
\end{quote}

\vspace{0.5em}\hrule\vspace{0.5em}

\paragraph{Iteration 80}
\begin{quote}
In a significant escalation of conflict, the Baloch Liberation Army (BLA) launched a bold attack on the Pakistani Frontier Corps headquarters in Panjgur, Balochistan, on February 10, inflicting heavy casualties and holding their position for nearly 40 hours. This attack represents the largest and most audacious by Baloch groups against Pakistani forces, whom they view as oppressive occupiers. The BLA, especially its elite Majeed Brigade, has a history of targeting both Pakistani and Chinese interests, as seen in attacks on projects related to the China Pakistan Economic Corridor (CPEC) and Chinese nationals. These actions stem from grievances over resource exploitation and marginalization of the Baloch people by both Pakistan and China. The recent surge in attacks follows a larger movement led by Baloch leaders like Jamaat-i-Islami's popular figure, amidst the Baloch community's growing hostility towards Sino-Pakistani cooperation.
\end{quote}

\vspace{0.5em}\hrule\vspace{0.5em}

\paragraph{Iteration 100}
\begin{quote}
Over a 40-hour period beginning on February 10, the Baloch Liberation Army (BLA) launched a bold attack on the Frontier Corps headquarters in Panjgur, Balochistan, marking one of their largest assaults against Pakistani forces. The BLA, representing Baloch groups who view the Pakistani government as a colonial and exploitative force, aims to dismantle perceived occupation and exploitation particularly linked to the China-Pakistan Economic Corridor (CPEC) projects which marginalize local Baloch people. Using their elite Majeed Brigade comprised of committed fidayeen, the BLA has a history of targeting both Pakistani and Chinese interests to express their grievances over resource deprivation and exploitative partnerships. Such violent campaigns have not only caused significant casualties but also unsettled China, which has responded by demanding increased security for its nationals in Pakistan. The attack fits into a pattern of recent escalations, reflecting the Baloch nationalists' deep-seated discontent.
\end{quote}

\vspace{0.5em}\hrule\vspace{0.5em}

\paragraph{Iteration 120}
\begin{quote}
In a recent escalation of violence, the Baloch Liberation Army (BLA) has launched a significant attack on the Pakistani Frontier Corps headquarters in Panjgur, Balochistan, maintaining control for 40 hours and inflicting heavy casualties on Pakistani forces. This marks one of the Baloch groups' most audacious attacks against what they perceive as an oppressive Pakistani regime, which alongside China, exploits Balochistan's resources while marginalizing its people. The surge in attacks coincides with growing discontent over the China Pakistan Economic Corridor (CPEC) that bypasses local employment. The BLA's elite Majeed Brigade, known for its previous high-profile assaults on Chinese interests, underscores the group's targeting of both Pakistani and Chinese influences in the region. Notable past attacks include assaults on Chinese workers and installations, showcasing the deep links Baloch nationalists have with regional geopolitics, further destabilizing Pakistan-China relations.
\end{quote}

\vspace{0.5em}
\hrule

\subsection{Screenshots of Annotation Interface}
\label{app:scr}
\clearpage
\input{figures/phase_1_review_article_fig}
\input{figures/phase_1_create_incident_fig}
\input{figures/phase_2_code_incident_fig}
\input{figures/phase_2_llm_fig}

%% file: figures/var.tex
\begin{figure*}
    \centering
    \includegraphics[width=\linewidth]{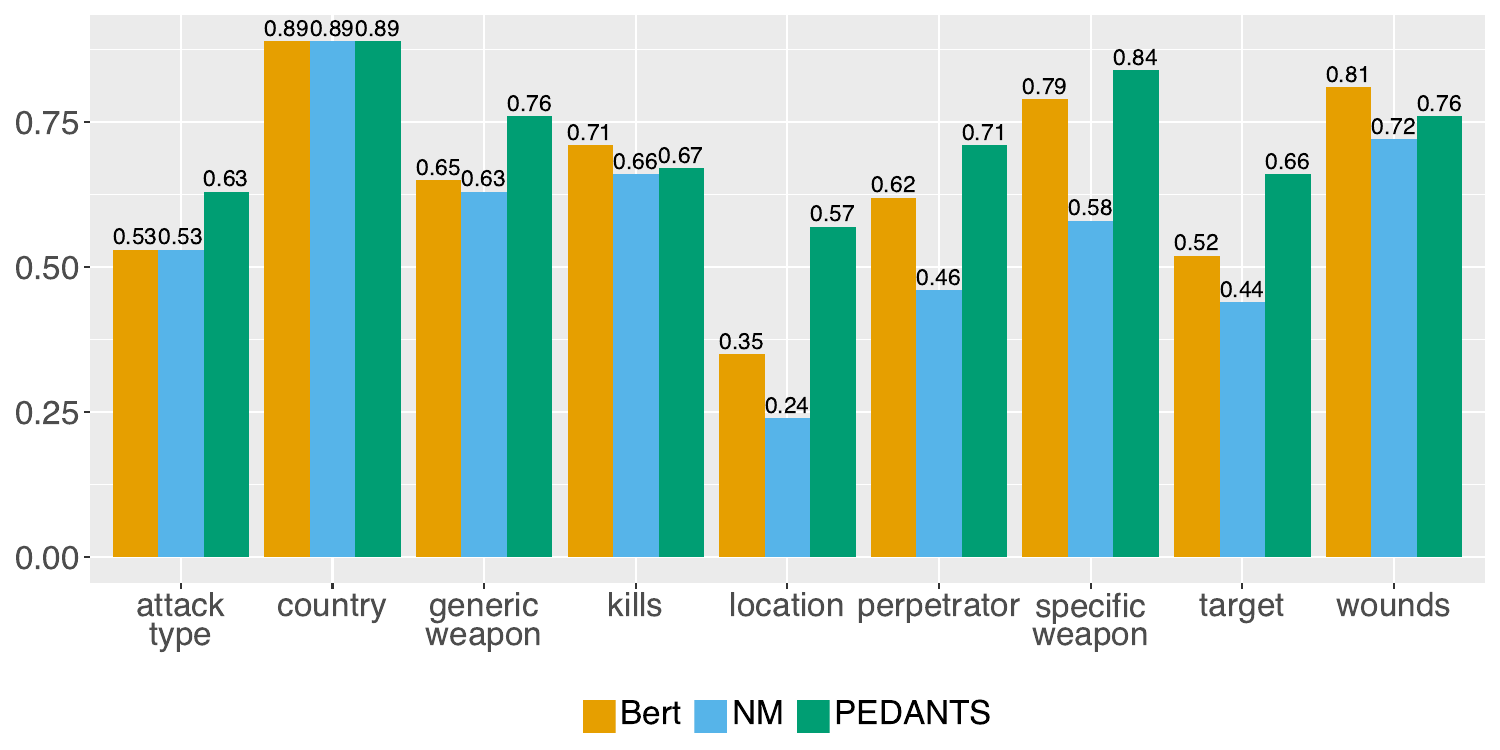}
    \caption{Agreement grouped by variable type. Human annotators agree more with extracted variables with higher degree of specificity. \texttt{Country} has over 90\% agreement. \textit{Generic attack type} and \textit{weapon type} also show high agreement. In comparison, low specificity variables like \textit{location} demonstrate low agreement with human judgment.}
    \label{fig:var}
\end{figure*}

%% file: tables/event_type_distr.tex
\begin{table}[h!]
\centering
\begin{tabular}{lr}
\hline
\textbf{Attack Type} & \textbf{Percentage} \\ \hline
Armed Assault & 30.66\% \\ 
Facility/Infrastructure & 13.68\% \\ 
Bombing/Explosion & 31.45\% \\ 
Hostage Taking (Kidnapping) & 7.55\% \\ 
Unarmed Assault & 1.10\% \\ 
N/A & 0.16\% \\ \hline
\end{tabular}
\caption{Attack Type Distribution}
\label{tab:event_type}
\end{table}

%% file: tables/weapon_type_distr.tex
\begin{table}[ht]
\centering
\begin{tabular}{lr}
\hline
\textbf{Generic Weapon} & \textbf{Percentage} \\ \hline
Explosives & 41.04\% \\ 
Firearms & 29.56\% \\ 
Incendiary & 8.96\% \\ 
Melee & 2.52\% \\ 
Other & 1.42\% \\ 
N/A & 30.82\% \\ \hline
\end{tabular}
\caption{Generic Weapon Type Distribution}
\label{tab:weapon_type}
\end{table}

%% file: figures/average.tex
\begin{figure}
    \centering
    \includegraphics[width=0.9\linewidth]{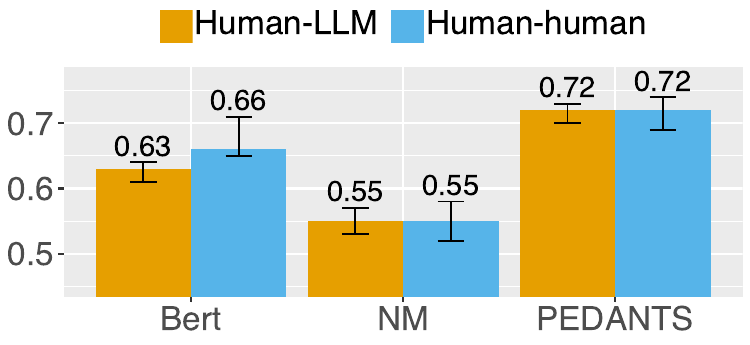} 
    \caption{
        In \cdeae{}, the percentage of agreement difference between human-human and human-\mm{} is not statistically significant, suggesting that \mm{}-coded variables provide human-level utility. On average, annotators and \textsc{gpt}-4o-mini agree 55\% using \textsc{nm}. \textsc{pedants} and \textsc{bem} show higher agreements.
    }
    \label{fig:avg}
\end{figure}

%% file: tables/agreement_acc_agg.tex
\begin{table}
\resizebox{\columnwidth}{!}{
\begin{tabular}{lrrr}
\toprule
\textbf{Model} & \textbf{\textsc{bem}} & \textbf{\textsc{em}} & \textbf{\textsc{pedant}} \\
\midrule
\textsc{claude-sonnet-4} & 0.47 & 0.40 & 0.50 \\
\textsc{gemini-2.5-flash} & 0.45 & 0.42 & 0.48 \\
\textsc{gemini-2.5-pro} & 0.45 & 0.39 & 0.50 \\
\textsc{gpt-4.1-mini} & 0.48 & 0.42 & 0.52 \\
\textsc{llama-3.3-70b-instruct} & 0.44 & 0.39 & 0.51 \\
\textsc{mistral-small-3.2-24b-instruct} & 0.46 & 0.40 & 0.46 \\
\textsc{mixtral-8x7b-instruct} & 0.39 & 0.28 & 0.46 \\
\textsc{qwen3-next-80b-a3b-instruct} & 0.48 & 0.43 & 0.51 \\
\bottomrule
\end{tabular}
}
\caption{Human-LLM agreement in CDEAE between expert annotators and different models. No model is significantly better at CDEAE.}
\label{tab:agg_acc_agreement}
\end{table}

%% file: tables/na_results.tex
\begin{table*}
\resizebox{\textwidth}{!}{
\begin{tabular}{lccccccc}
\toprule
\textbf{Model} & \textbf{Precision} & \textbf{Recall} & \textbf{F1-Score} & \textbf{TP} & \textbf{FP} & \textbf{FN} & \textbf{TN} \\
\midrule
\textsc{claude-sonnet-4} & 0.51 & 0.27 & 0.35 & 193 & 189 & 533 & 1947 \\
\textsc{gemini-2.5-flash} & 0.43 & \textbf{0.39} & \textbf{0.41} & 284 & 377 & 442 & 1759 \\
\textsc{gemini-2.5-pro}& 0.44 & 0.24 & 0.31 & 172 & 217 & 554 & 1919 \\
\textsc{gpt-4.1-mini} & 0.53 & 0.20 & 0.30 & 148 & 129 & 578 & 2007 \\
\textsc{llama-3.3-70b-instruct}  & 0.51 & 0.28 & 0.36 & 204 & 198 & 522 & 1938 \\
\textsc{mistral-small-3.2-24b-instruct} & 0.48 & 0.36 & \textbf{0.41} & 262 & 289 & 464 & 1847 \\
\textsc{mixtral-8x7b-instruct} & \textbf{0.62} & 0.03 & 0.06 & 24 & 15 & 702 & 2121 \\
\textsc{qwen3-next-80b-a3b-instruct} & 0.52 & 0.25 & 0.34 & 180 & 167 & 546 & 1969 \\
\bottomrule
\end{tabular}
}
\caption{Model accuracies in identifying variables that are not available in documents. \textsc{mixtral-8x7b-instruct} has the highest precision because it almost always code variables even when they are not available. Other models do not differ significantly from the general population.}
\label{tab:na_results}
\end{table*}

%% file: figures/phase_1_review_article_fig.tex
\begin{figure*}[p]
    \centering
    \includegraphics[width=0.75\linewidth]{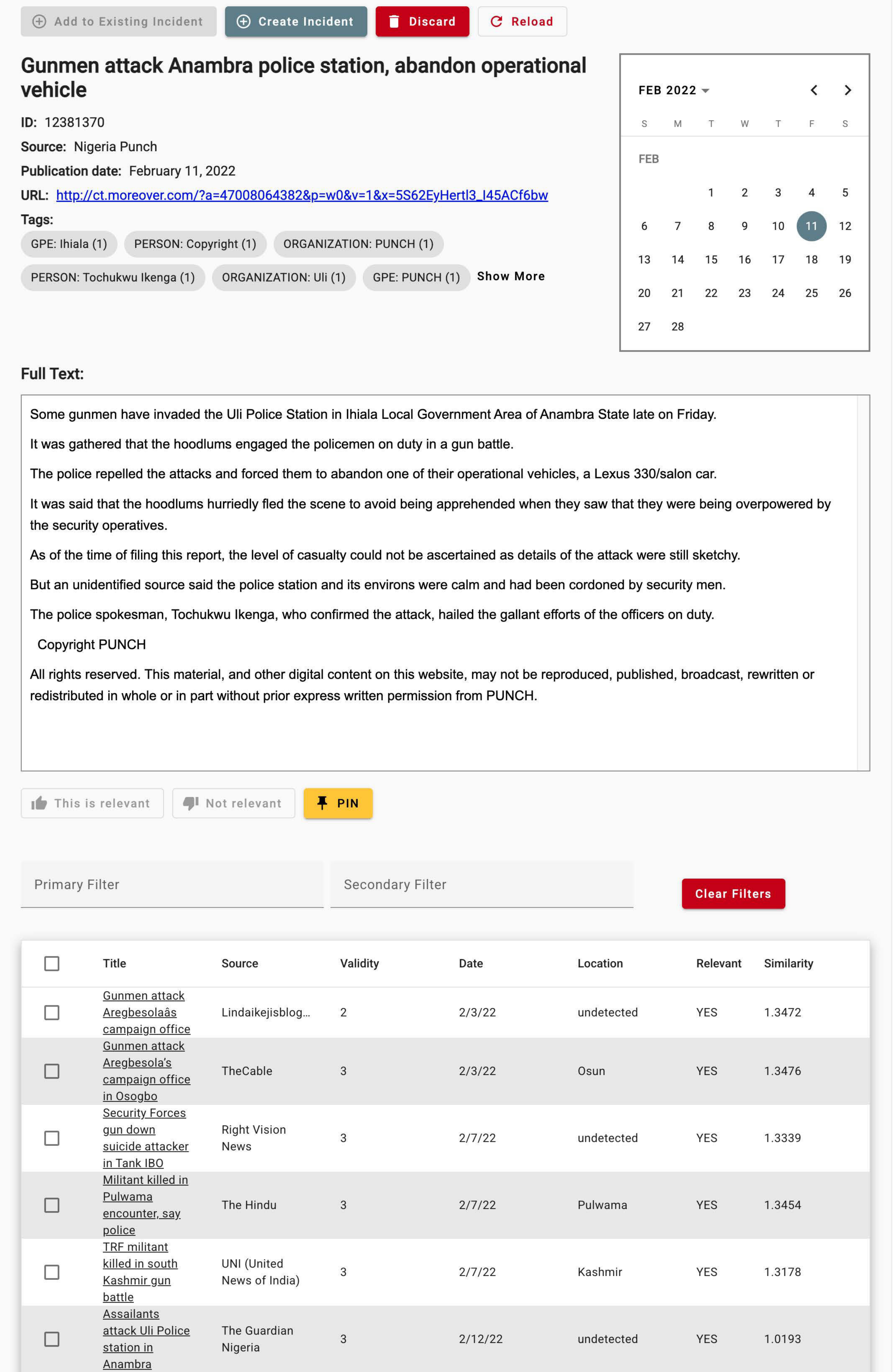} 
    \caption{An example of CDCR interface. Annotators see a document along with similar documents retrieved by \abr{tf-idf}}
    \label{fig:phase_1_review_article}
\end{figure*}

%% file: figures/phase_1_create_incident_fig.tex
\begin{figure*}[p]
    \centering
    \includegraphics[width=0.75\linewidth]{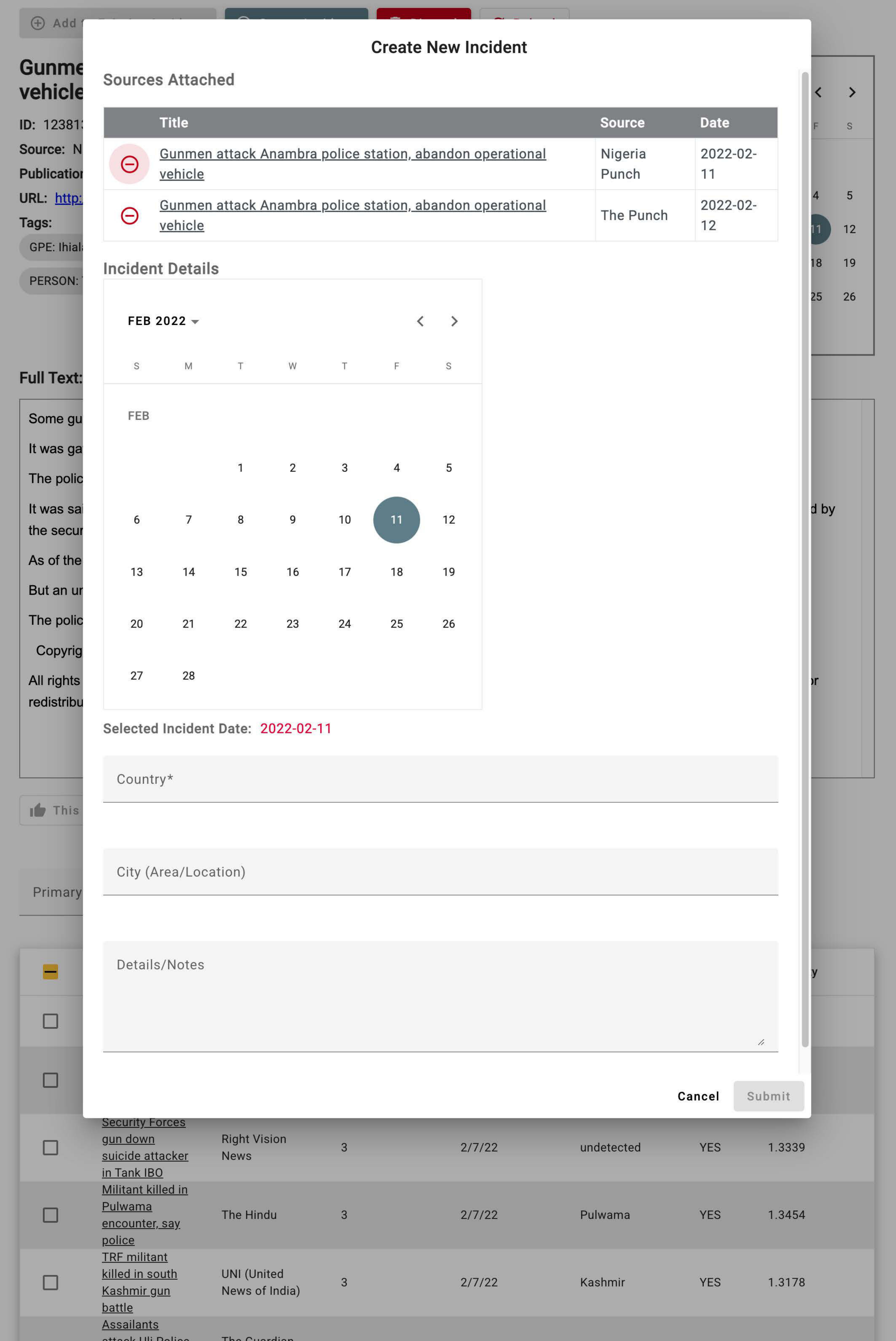} 
    \caption{In CDCR, once an annotator finds all relevant documents, they create the events and fill in the preliminary variables for the second phase\textendash extraction.}
    \label{fig:phase_1_create_incident}
\end{figure*}

%% file: figures/phase_2_code_incident_fig.tex
\begin{figure*}[p]
    \centering
    \includegraphics[width=0.75\linewidth]{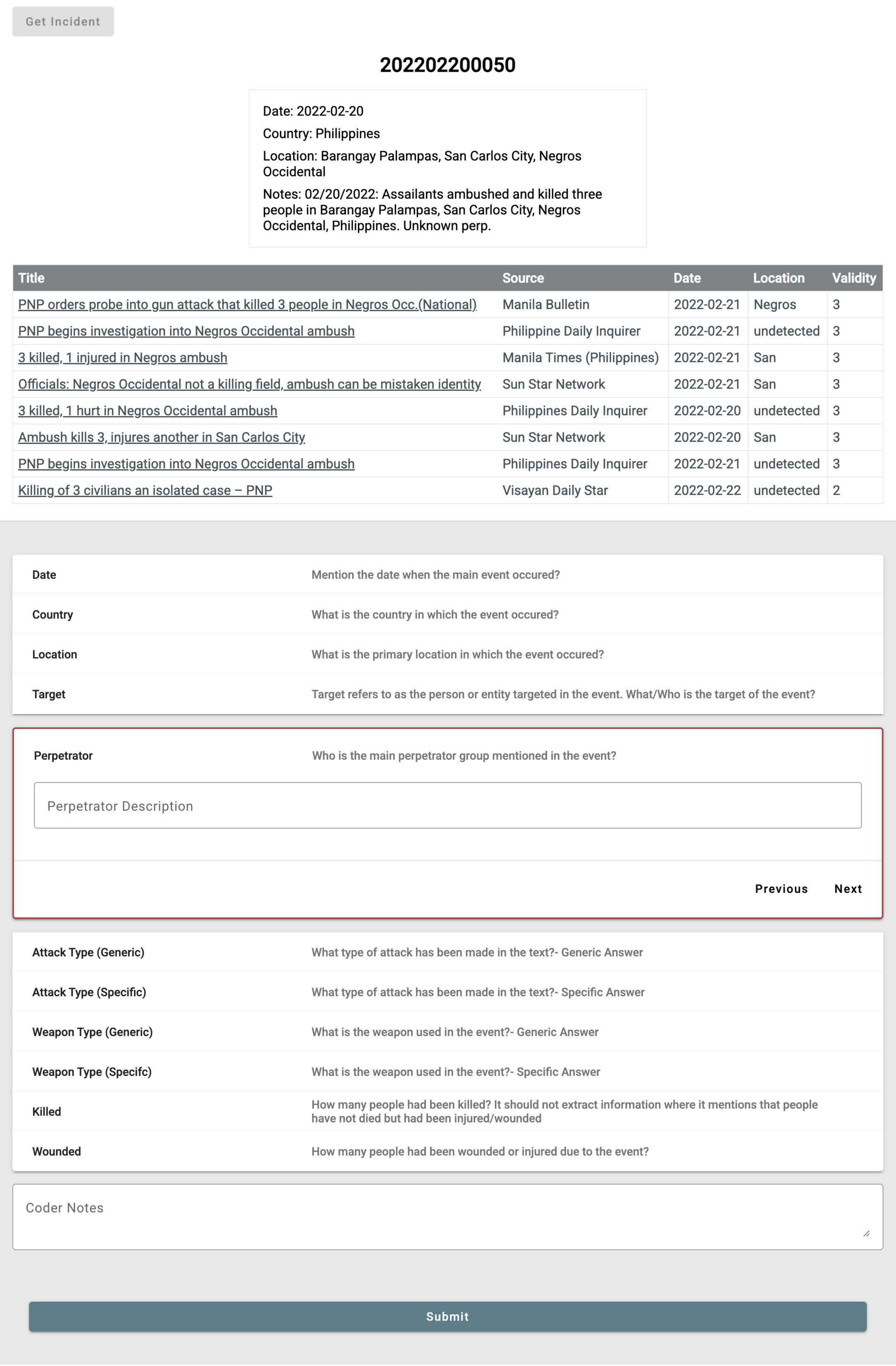} 
    \caption{Existing pipeline where annotators code the variables without LLM assistance.}
    \label{fig:phase_2_code_incident}
\end{figure*}

%% file: figures/phase_2_llm_fig.tex
\begin{figure*}[p]
    \centering
    \includegraphics[width=0.75\linewidth]{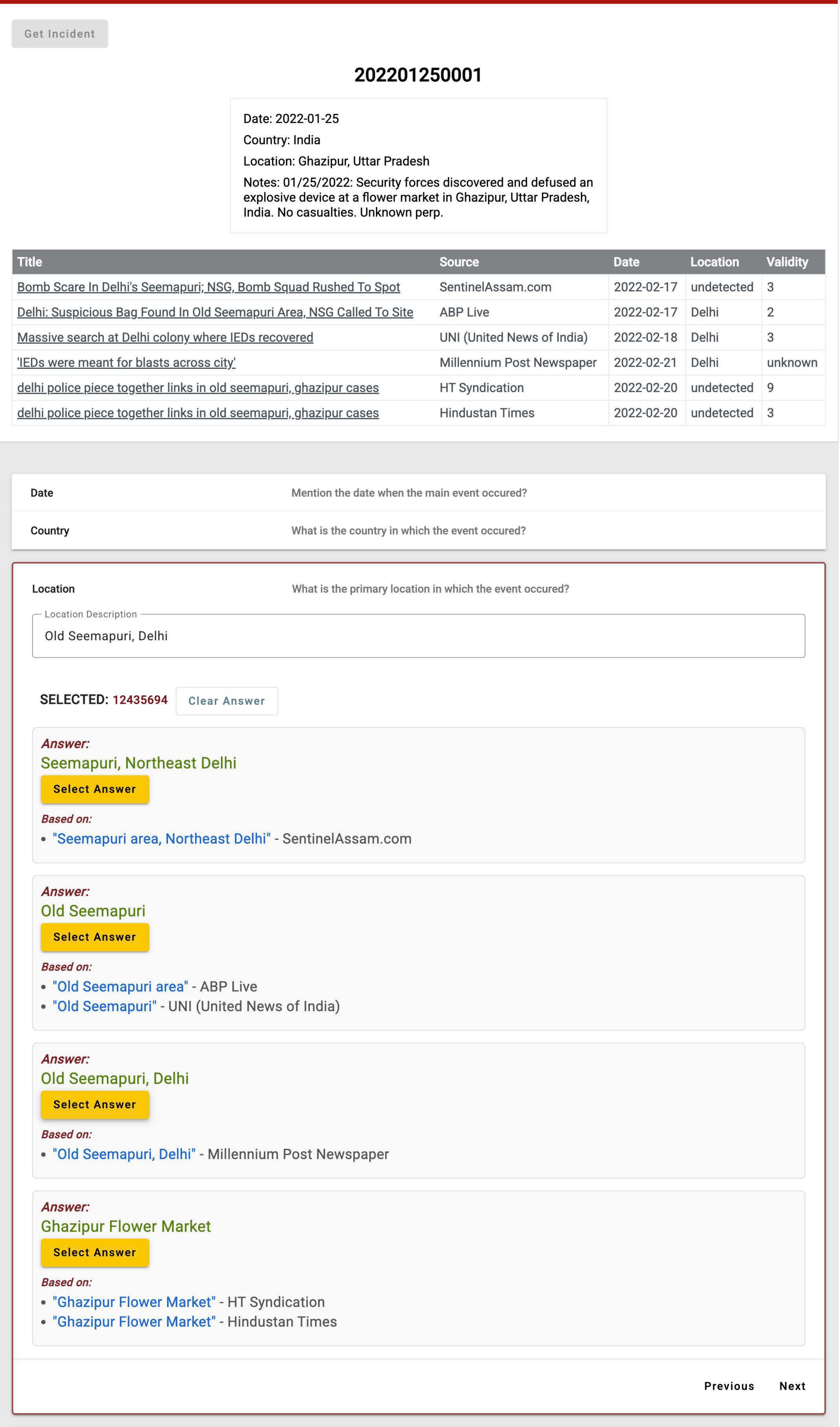} 
    \caption{Interface where annotators see the LLM-extracted variables.}
    \label{fig:phase_2_llm}
\end{figure*}